\documentclass[review]{elsarticle}
\usepackage[margin=1in]{geometry}
\usepackage{amsmath}
\usepackage{amssymb}
\usepackage{hyperref}
\usepackage{doi}
\usepackage[numbers]{natbib}
\usepackage[utf8]{inputenc}
\usepackage{tikz}
\usepackage{pgfplots}
\usepackage{booktabs}
\usepackage{subcaption}
\usepackage{multirow} 
\usepackage{url}
\usetikzlibrary{shapes.geometric, arrows.meta, positioning}

\begin{document}

\begin{frontmatter}

\title{Optimized Hybrid Feature Engineering for Resource-Efficient Arrhythmia Detection in ECG Signals: An Optimization Framework}

\author[1]{Moirangthem Tiken Singh\corref{cor1}}
\ead{tiken.m@dibru.ac.in}

\author[2]{Manibhushan Yaikhom}
\ead{mbyaikhom@gmail.com}

\affiliation[1]{organization={Department of Computer Science and Engineering, Dibrugarh University Institute of Engineering and Technology (DUIET), Dibrugarh University},
            addressline={Dibrugarh, Assam}, 
            country={India}}

\affiliation[2]{organization={Department of Physical Medicine and Rehabilitation, Regional Institute of Medical Sciences},
            addressline={Imphal, Manipur}, 
            country={India}}

\cortext[cor1]{Corresponding author}
% \begin{document}
	
% 	\title{Optimized Hybrid Feature Engineering for Resource-Efficient Arrhythmia Detection in ECG Signals: An Optimization Framework}
% 	\author{
% Moirangthem Tiken Singh \\
% Department of Computer Science and Engineering \\
% Dibrugarh University Institute of Engineering and Technology (DUIET) \\
% Dibrugarh University, Dibrugarh, Assam, India \\
% tiken.m@dibru.ac.in
% }
% 	\date{}
% 	\maketitle
\begin{abstract}
Cardiovascular diseases, particularly arrhythmias, remain a leading global cause of mortality, necessitating continuous monitoring via the Internet of Medical Things (IoMT). However, state-of-the-art deep learning approaches often impose prohibitive computational overheads, rendering them unsuitable for resource-constrained edge devices. This study proposes a resource-efficient, data-centric framework that prioritizes feature engineering over complexity. Our optimized pipeline makes the complex, high-dimensional arrhythmia data linearly separable. This is achieved by integrating time-frequency wavelet decompositions with graph-theoretic structural descriptors, such as PageRank centrality. This hybrid feature space, combining wavelet decompositions and graph-theoretic descriptors, is then refined using mutual information and recursive elimination, enabling interpretable, ultra-lightweight linear classifiers. Validation on the MIT-BIH and INCART datasets yields 98.44\% diagnostic accuracy with an 8.54 KB model footprint. The system achieves 0.46 $\mu$s classification inference latency within a 52 ms per-beat pipeline, ensuring real-time operation. These outcomes provide an order-of-magnitude efficiency gain over compressed models, such as KD-Light (25 KB, 96.32\% accuracy), advancing battery-less cardiac sensors.
\end{abstract}


\end{frontmatter}

\section{Introduction}
Cardiovascular diseases (CVDs) remain the predominant global cause of mortality, contributing to an estimated 19.2 million deaths in 2023, a marked increase from 13.1 million in 1990, underscoring the escalating public health burden \cite{ACC2025_CVD_GlobalDeaths}. Arrhythmias, in particular, serve as critical harbingers of severe complications such as ischemic stroke and sudden cardiac arrest \cite{attia2019screening}. This upward trajectory in prevalence, driven by aging populations and rising non-communicable disease risks, amplifies the need for scalable diagnostic solutions. The advent of the Internet of Medical Things (IoMT) offers a transformative opportunity to evolve cardiac monitoring from episodic Holter evaluations to persistent, real-time oversight~\cite{dowdeswell2023healthcare}. However, realizing this potential is impeded by a fundamental asymmetry: while sensor hardware has progressed toward greater compactness and energy optimization, the concomitant algorithms for signal interpretation have become increasingly computationally intensive, thereby constraining deployment on resource-limited platforms~\cite{dwivedi2022potential}. 

An efficacious edge-based monitoring system requires the fulfillment of three key criteria: diagnostic acuity comparable to expert clinical judgment, energy consumption commensurate with microcontroller constraints, and interpretability sufficient to generate clinical trust. Historically, research has privileged acuity, culminating in a model-centric framework dominated by advanced deep neural architectures, including Convolutional Neural Networks (CNNs) and Transformers \cite{Alghieth2025DeepECGNet}. While yielding elevated accuracy metrics, these models function as inscrutable systems that impose significant power burdens on edge infrastructure, often exceeding 100 mW for inference alone, and omit the transparent decision mechanisms that are imperative for medical validation. Quantitative analyses have revealed that such architectures can inflate latency by factors of $10^3$ to $10^4$ relative to linear alternatives, exacerbating battery drain in prolonged monitoring scenarios \cite{lin2024discrete}. Conventional remediation strategies, such as model compression via quantization, often engender trade-offs in performance, with accuracy degradations of 1-5\% commonly observed, and perpetuate the interpretability deficit, an attenuated opaque model retains its fundamental intransparency \cite{8768224}, underscoring the need for an alternative paradigm that intrinsically addresses these challenges. Therefore, to overcome these persistent limitations and address the critical need for both efficiency and interpretability, our study posits that a fundamental re-evaluation of the approach is necessary, shifting the emphasis from architectural complexity to feature-centric refinement. 

We hypothesize that the multifaceted, high-dimensional arrhythmia data manifold mapping to a linearly separable representation through systematic, domain-grounded transformations, thereby facilitating the use of ultra-lightweight and interpretable linear classifiers. This involves engineering a hybrid feature ensemble that amalgamates wavelet-based time-frequency decompositions \cite{daubechies1992ten} with graph-theoretic structural metrics \cite{andayeshgar2024arrhythmia}, thereby reducing the classification manifold to a form that can be handled by rudimentary linear solvers while preserving separability metrics. 

To evaluate this hypothesis and guide the investigation, this study addresses the following three principal research questions: 
\begin{enumerate} 
\item[(1)] To what extent does hybrid feature engineering enable linear separability in arrhythmia classification, facilitating high-accuracy diagnostics with low computational overhead? 
\item[(2)] What are the quantifiable trade-offs in performance, model size, and inference latency compared to deep learning paradigms? 
\item[(3)] How robust is the framework across diverse datasets, including single-lead and multi-lead configurations? 
\end{enumerate} 

In addressing these questions, the proposed framework demonstrates robust performance attributes, including high accuracy, computational efficiency and inherent explicability. Validation across benchmarks, such as MIT-BIH~\cite{moody2001impact} and INCART~\cite{incartdb}, confirms that our interpretable linear models achieve clinical-grade diagnostics by effectively discriminating between morphologies, paving the way for trustworthy, low-latency edge AI.

\section{Literature Survey}
The automated detection of cardiac arrhythmias has undergone a transformative evolution, shifting from heuristic-based signal processing to high-capacity deep learning architectures. This section categorizes the current state-of-the-art across four thematic domains: deep learning paradigms, model compression for the edge, hybrid feature engineering, and the recent transition toward lightweight linear separability.

% \subsection{Deep Learning and the Complexity-Performance Paradigm}
Recent breakthroughs in arrhythmia classification have been dominated by Convolutional Neural Networks (CNNs) and Recurrent Neural Networks (RNNs). Studies utilizing the MIT-BIH database \cite{mitdb} have demonstrated that 1D-CNNs can achieve accuracies exceeding 99\% by learning hierarchical morphological features directly from raw ECG signals \cite{hannun2019cardiologist, kiranyaz2015real}. Manal Alghieth \cite{Alghieth2025DeepECGNet} recently demonstrated that Transformer-based models could further improve sensitivity for rare ectopic beats by leveraging self-attention mechanisms to capture long-range temporal dependencies. Similarly, the integration of Gated Recurrent Units (GRUs) has proved effective in modeling the sequential nature of long-term Holter recordings \cite{li2024gru}. A comprehensive review by Reshad et al. \cite{reshad2025deep} evaluated 30 deep learning studies from 2022-2025, highlighting CNN-RNN hybrids achieving up to 99.93\% accuracy, though at the cost of computational intensity. Kim et al. \cite{kim2025} proposed a hybrid CNN-transformer model for the detection of arrhythmias without the identification of the R-peak, achieving 97.8\% precision in large datasets. However, these models often require millions of trainable parameters, necessitating high-performance GPUs and incurring substantial inference latencies, which limit their utility in edge devices \cite{alghieth2025deepecg}. While these deep learning models achieve impressive accuracy, their inherent computational intensity and large parameter counts remain a significant bottleneck for deployment on resource-constrained edge devices, necessitating alternative paradigms.

% \subsection{Model Compression and Lightweight Architectures for the Edge}
To bridge the gap between clinical-grade accuracy and edge feasibility, researchers have explored various model compression techniques, such as 8-bit quantization and knowledge distillation. Authors in \cite{9630348} utilized TensorFlow Lite to compress ResNet and MobileNet architectures to approximately 70 KB, maintaining an accuracy of 93.7\%. Similarly, the KD-Light framework \cite{s24247896} employed knowledge distillation to produce a student model 1242$\times$ smaller than its teacher, though it resulted in a notable performance drop to 96.32\%. Other approaches focus on hardware-co-optimization, using FPGA-based accelerators~\cite{muthuramalingam2008neural} to reduce the power consumption of quantized Deep Convolutional Neural Networks (DCNNs) \cite{rawal2023hardware}. An et al. \cite{MENG2022102236} developed the Lightweight Fussing Transformer (LFT) model, an approach designed to address the
challenges of processing noisy and dynamic ECG signals from wearable devices. The proposed model achieved 99.32\% with parameter reduction of 72\% compared to
traditional self-attention architectures. Jiang et al. \cite{Jiang2024} introduced DCETEN, a Transformer-based lightweight network with pruning, yielding 99.84\% accuracy on MIT-BIH with low resource demands. These methods illustrate a fundamental Pareto trade-off: extreme compression often sacrifices diagnostic precision, particularly in distinguishing between morphologically similar classes like Supraventricular and Normal beats \cite{8768224}.

% \subsection{Hybrid Feature Engineering and Compressive Sensing}
While model compression offers some mitigation, a parallel branch of research suggests that the complexity of the classifier can be drastically reduced if the underlying data manifold is effectively unrolled using domain-informed feature engineering, potentially circumventing the performance trade-offs inherent in model compression. Hybrid approaches combining discrete wavelet transforms (DWT) for spectral analysis with statistical moments (skewness, kurtosis) have shown promise in creating robust feature representations \cite{10263710, martis2013ecg, ghaffari2008new}. Recent work has also introduced graph-theoretic descriptors, such as PageRank centrality, to quantify beat-to-beat similarity, treating ECG records as structural networks rather than mere temporal sequences \cite{andayeshgar2024arrhythmia}. Bian et al. \cite{9433552} demonstrated that classifying signals in the compressive sensing domain could reduce energy consumption, though at the cost of reduced accuracy. Furthermore, shallow state-space models have emerged as a noise-resilient alternative for edge monitoring, albeit with limited success in multi-class classification scenarios \cite{Huang2024S4DECG}. Mechichi and Benzarti \cite{Mechichi2025} proposed a CNN-BiLSTM-attention hybrid for ECG classification, achieving 99.20\% accuracy with SMOTE balancing. Varalakshmi and Sankaran \cite{Varalakshmi2023} developed a BiLSTM-Random Forest model with 98.84\% accuracy, incorporating PCA for feature reduction. Raunsay et al. \cite{11211275} advanced a Hybrid Transformer BERT-CNN, reaching 99\% on MIT-BIH without R-peak detection.

% \subsection{Research Gap and Contributions}
While existing literature provides a robust foundation for either high-accuracy deep learning models or ultra-low-power compressive sensing, there remains a critical gap for a framework that offers clinical-grade accuracy ($>98\%$) with sub-10 KB memory footprints. Most current lightweight models still rely on shallow neural networks that, despite quantization, retain significant computational depth and latency \cite{Lee2025ArrhythmiaCNN, liu2023tinyml}. Pósfay et al. \cite{posfay2025} introduced LLT-ECG, a linear law-based feature extraction achieving near-linear separability with 96.4\% accuracy using simple classifiers. This study addresses this gap by proposing a hybrid feature augmentation strategy. Unlike the model-centric SOTA, our approach shifts the locus of complexity from the classifier to the feature space, utilizing a 2-lead configuration to enable ultra-fast, microsecond-range inference via linear solvers. This strategy aims to prove that rigorous feature engineering can achieve the diagnostic yield of deep networks while remaining compatible with the most stringent hardware constraints of next-generation cardiac edge devices.

\section{Methodology}

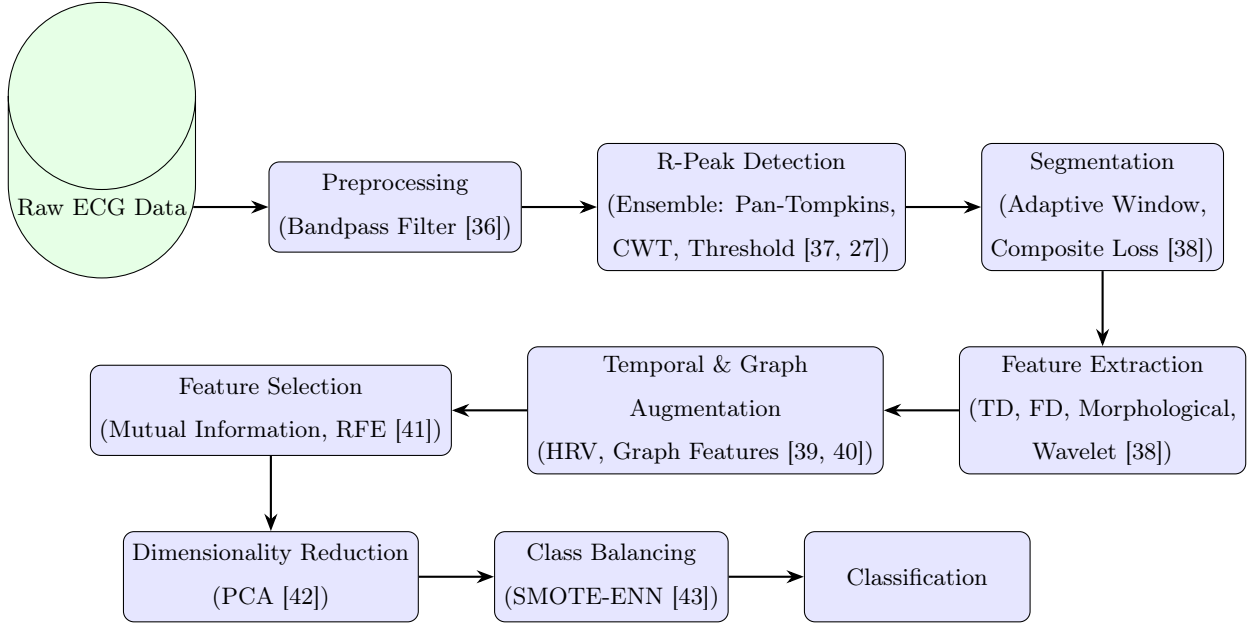
\begin{figure*}[h!]
\centering
\begin{tikzpicture}[
    process/.style={rectangle, draw, rounded corners, minimum height=1.2cm, minimum width=3cm, align=center, fill=blue!10},
    arrow/.style={-Stealth, thick},
    data/.style={cylinder, shape border rotate=90, draw, minimum height=2cm, minimum width=1.5cm, fill=green!10},
    font=\small
]
% Nodes
\node[data] (data) at (0,0) {Raw ECG Data};
\node[process, right=1cm of data] (preprocess) {Preprocessing\\(Bandpass Filter~\cite{selesnick1998generalized})};
\node[process, right=1cm of preprocess] (rpeak) {R-Peak Detection\\(Ensemble: Pan-Tompkins,\\CWT, Threshold~\cite{pan1985real,ghaffari2008new})};
\node[process, right=1cm of rpeak] (segment) {Segmentation\\(Adaptive Window,\\Composite Loss~\cite{romero2020delineation})};
\node[process, below=1cm of segment] (feature) {Feature Extraction\\(TD, FD, Morphological,\\Wavelet~\cite{romero2020delineation})};
\node[process, left=1cm of feature] (augment) {Temporal \& Graph\\Augmentation\\(HRV, Graph Features~\cite{welch1967use,page1999pagerank})};
\node[process, left=1cm of augment] (select) {Feature Selection\\(Mutual Information, RFE~\cite{darst2018using})};
\node[process, below=1cm of select] (reduce) {Dimensionality Reduction\\(PCA~\cite{jolliffe2016principal})};
\node[process, right=1cm of reduce] (balance) {Class Balancing\\(SMOTE-ENN~\cite{batista2004study})};
\node[process, right=1cm of balance] (classify) {Classification};

% Arrows
\draw[arrow] (data) -- (preprocess);
\draw[arrow] (preprocess) -- (rpeak);
\draw[arrow] (rpeak) -- (segment);
\draw[arrow] (segment) -- (feature);
\draw[arrow] (feature) -- (augment);
\draw[arrow] (augment) -- (select);
\draw[arrow] (select) -- (reduce);
\draw[arrow] (reduce) -- (balance);
\draw[arrow] (balance) -- (classify);
\end{tikzpicture}
\caption{ECG Classification Pipeline for Arrhythmia Detection}
\label{fig:pipeline}
\end{figure*}

The ECG classification pipeline, illustrated in Figure~\ref{fig:pipeline}, is engineered to establish a robust supervised learning framework for arrhythmia detection. The process transforms raw multichannel ECG recordings into a high-dimensional discriminative feature matrix through a sequence of optimized stages: preprocessing, R-peak detection, adaptive segmentation, feature extraction, and augmentation. These stages are succeeded by feature selection, dimensionality reduction, and class balancing to prepare the data for classification. Each component is formulated as an optimization problem designed to maximize signal quality, information retention, or classification accuracy, as detailed in the subsequent sections.

The initial stage focuses on attenuating baseline wander and high-frequency noise, which can obscure QRS complexes and compromise detection accuracy. To ensure signal fidelity, we apply a bandpass Butterworth filter designed to minimize out-of-band energy while strictly preserving QRS morphology. The passband is aligned with the standard ECG frequency spectrum (0.5–40 Hz) to capture clinically relevant components while rejecting artifacts~\citep{selesnick1998generalized}.

\begin{figure}[h!]
    \centering
    \includegraphics[width=0.7\linewidth]{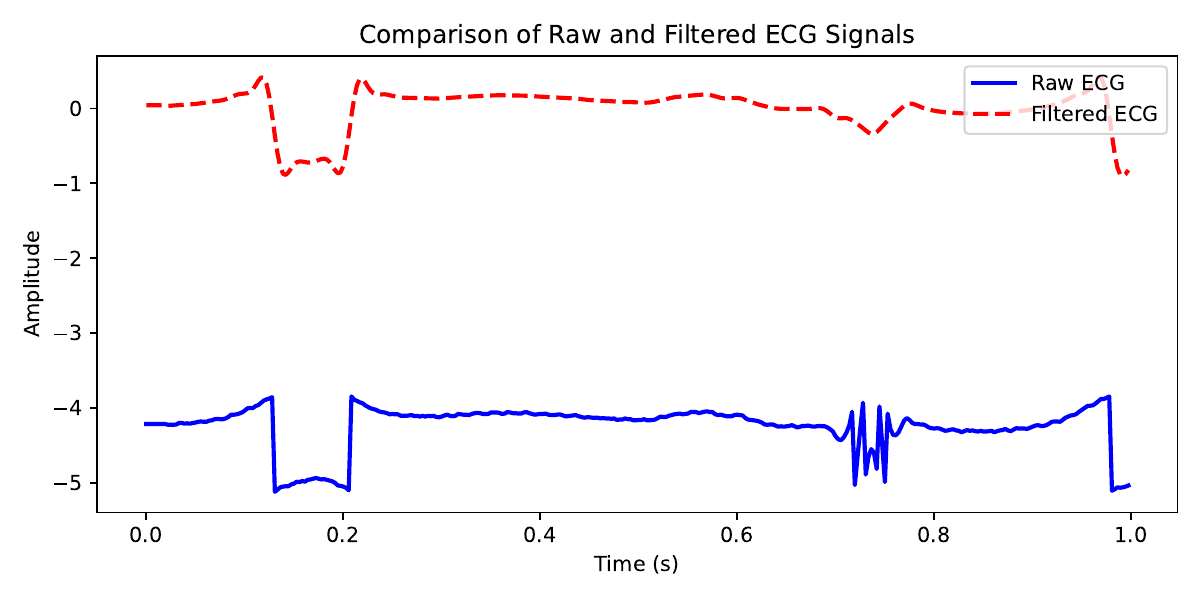}
    \caption{Comparison of raw and preprocessed ECG signals, illustrating the effect of filtering and noise removal}
    \label{fig:preprocessing}
\end{figure}

For digital implementation, discrete-time coefficients are derived from the analog prototype via bilinear transformation to prevent frequency warping~\cite{oppenheim1999discrete}. The filter is realized as a cascade of second-order sections (biquads) to enhance numerical stability, governed by the transfer function:
\begin{equation}
    H(z) = \prod_{k} \frac{b_{0k} + b_{1k}z^{-1} + b_{2k}z^{-2}}{a_{0k} + a_{1k}z^{-1} + a_{2k}z^{-2}}
\end{equation}
where coefficients $b_{ik}$ and $a_{ik}$ are optimized for minimal passband ripple. Zero-phase filtering is achieved through forward and reverse application, eliminating phase distortion—a critical requirement for preserving the timing of temporal features. Figure~\ref{fig:preprocessing} demonstrates the efficacy of this stage.

\begin{figure}[h!]
\centering
\includegraphics[width=0.7\linewidth]{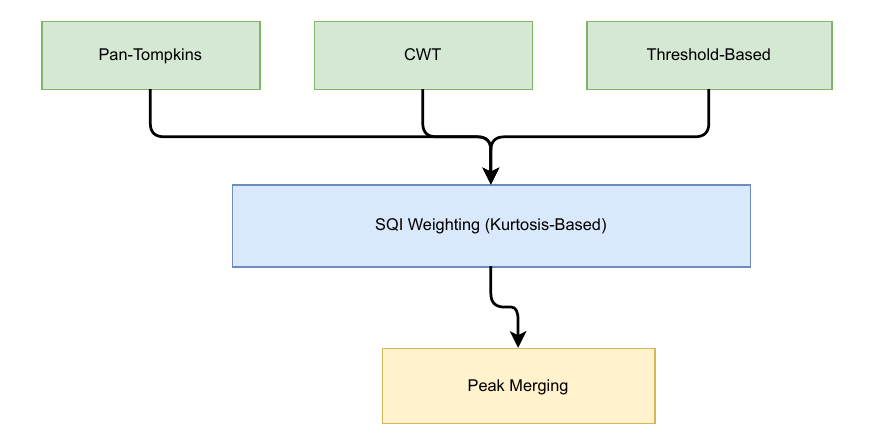}
\caption{Ensemble Approach for R-Peak Detection with SQI Weighting}
\label{fig:rpeak_ensemble}
\end{figure}

With signal fidelity restored, the pipeline proceeds to R-peak detection, a prerequisite for accurately localizing cardiac cycles. We employ an ensemble approach combining Pan-Tompkins (differentiation and integration), CWT (Mexican hat wavelet energy maximization), and adaptive thresholding~\citep{pan1985real,ghaffari2008new}. As shown in Figure~\ref{fig:rpeak_ensemble}, the detectors are weighted by a Signal Quality Index (SQI) based on kurtosis, prioritizing high-quality inputs. Detected peaks are merged within a physiological refractory tolerance to eliminate duplicates.

Subsequent segmentation delineates individual beats. To address RR-interval variability caused by arrhythmias, we utilize an adaptive windowing technique that minimizes a composite loss function per R-peak:
\begin{equation}
    L(\alpha, \beta) = w_e H + w_s (1 - \text{SNR}) + w_g (1 - E_r) + P_l
\end{equation}
where $H$ represents Shannon entropy, SNR is the QRS-band energy ratio, $E_r$ is the QRS-to-total energy ratio, and $P_l$ is a penalty for inappropriate segment lengths~\citep{romero2020delineation}. Outliers are pruned via interquartile range filtering, and valid segments are resampled to a fixed length using polyphase resampling.

To discern subtle inter-class differences, we extract a multidimensional feature set encompassing time-domain (statistical moments, zero-crossing rates), frequency-domain (spectral entropy, band power), and morphological attributes~\cite{elhaj2016arrhythmia}. Frequency features are derived via the discrete Fourier transform:
\begin{equation}
    X(k) = \sum_{n=0}^{N-1} x(n) e^{-j 2\pi k n / N}
\end{equation}
where $x(n)$ denotes the discrete time-domain input sequence of length $N$, $n$ is the sample index, and $X(k)$ represents the spectral component at frequency bin $k$.

Morphological features include QRS duration, T-wave amplitude, and ST deviation, derived from CWT-based delineation. Additionally, wavelet packet energies are computed using Daubechies decomposition.

Since arrhythmias often manifest as sequence-dependent patterns rather than isolated beat anomalies, we apply temporal and graph augmentation to model inter-beat dynamics. Temporal augmentation computes Heart Rate Variability (HRV) metrics in two domains:
\begin{enumerate}
    \item Time-Domain: SDNN (standard deviation of NN intervals) and pNN50 (percentage of intervals differing by $>50$ms) are calculated directly from RR-interval statistics to quantify overall variability.
    \item Frequency-Domain: The LF/HF ratio is derived using Welch’s method to estimate the Power Spectral Density (PSD) of the RR series:
    \begin{equation}
        P(f) = \frac{1}{M} \sum_{m=0}^{M-1} \left| X_m(f) \right|^2
    \end{equation}
    where $M$ is the number of overlapping segments and $X_m(f)$ is the windowed Fourier transform of the $m$-th segment. This reduces noise variance compared to a single periodogram.
\end{enumerate}
\begin{figure}
    \centering
    \includegraphics[width=0.5\linewidth]{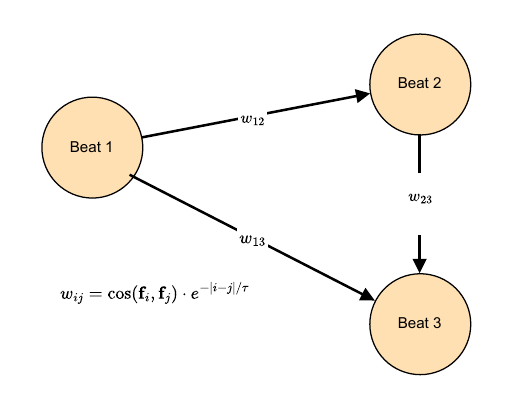}
    \caption{Directed Graph for graph augmentation.}
    \label{fig:placeholder}
\end{figure}
Complementing the temporal analysis, graph augmentation constructs a directed graph where nodes represent beats and edges encode similarity and temporal proximity. The edge weights $w_{ij}$ between beats $i$ and $j$ are calculated as:
\begin{equation}
    w_{ij} = \cos(f_i, f_j) \cdot e^{-|i-j|/\tau}
\end{equation}
where $f_i$ and $f_j$ denote the extracted feature vectors (morphological and spectral attributes) for the $i$-th and $j$-th beats, respectively, and $\tau$ is a temporal decay constant. Graph-theoretic features are then extracted to quantify the structural topology of the beat sequence:

1. PageRank Centrality: This metric identifies the most ``representative'' beats in the sequence. It is calculated iteratively using the power method:
\begin{equation}
    PR = (1-d) 1 + d A D^{-1} PR
\end{equation}
Here, $PR$ is the centrality vector, $d$ is the damping factor (typically 0.85), $1$ is a vector of ones representing the teleportation probability (ensuring exploration of the whole graph), $A$ is the weighted adjacency matrix, and $D^{-1}$ normalizes the columns to create a stochastic transition matrix. High PageRank values indicate beats that share a strong morphological similarity with many other influential beats, helping the model distinguish consistent rhythms from artifacts.

2. Clustering Coefficient: To measure local morphological stability, we compute the weighted clustering coefficient for each node $i$~\cite{barrat2004architecture}:
\begin{equation}
    C_i = \frac{1}{s_i (k_i - 1)} \sum_{j,h} \frac{w_{ij} + w_{ih}}{2} a_{ij} a_{ih} a_{jh}
\end{equation}
where $k_i$ is the degree of node $i$, $s_i$ is the node strength (sum of weights), $w_{ij}$ represents the edge weights, and $a_{ij}$ is the binary adjacency status (1 if connected, 0 otherwise). This quantifies the likelihood that two neighbors of a beat are also similar to each other, providing a robust feature for detecting irregular arrhythmias, such as Atrial Fibrillation, where morphological consistency is disrupted.

To manage the high dimensionality of the augmented feature space while preserving both granular details and global variance, we employ a hybrid optimization process. First, feature selection reduces redundancy by thresholding Mutual Information (MI)~\cite{cover2006elements}:
\begin{equation}
    I(X; Y) = \sum_{x \in X} \sum_{y \in Y} p(x,y) \log \frac{p(x,y)}{p(x)p(y)}
\end{equation}
This is complemented by Recursive Feature Elimination (RFE), which identifies a subset of the most significant features by iteratively minimizing the classification error:
\begin{equation}
    \min \sum (y - \hat{y})^2
\end{equation}

Second, we utilize Principal Component Analysis (PCA) as a feature enhancement technique rather than for strict dimensionality reduction. PCA is applied to the RFE-selected subset to extract the top principal components maximizing explained variance:
\begin{equation}
    \max \sum \lambda_i
\end{equation}
where $\lambda_i$ represents the eigenvalues of the covariance matrix. These components are then concatenated with the original augmented feature matrix. This strategy yields an expanded feature vector that combines specific, interpretable local attributes with compact, denoised global variance descriptors.

Finally, to address the class imbalance inherent in arrhythmia datasets, we utilize SMOTE-ENN to synthesize minority samples and clean decision boundaries~\citep{batista2004study}. The balanced, enriched feature matrix trains a set of interpretable linear classifiers (e.g., Logistic Regression, Linear SVM~\cite{cortes1995support}). The optimal parameters $\theta^*$ are obtained by minimizing the regularized empirical risk~\cite{bishop2006pattern}:
\begin{equation}
\theta^* = \arg\min_\theta \frac{1}{N} \sum_{i=1}^N \ell\bigl(f_\theta(\mathbf{x}_i), y_i\bigr) + \alpha R(\theta)
\end{equation}
where $\ell(\cdot)$ denotes the loss function, $R(\theta)$ is the regularization term, and $\alpha$ controls the regularization strength. This formulation ensures equitable performance across all arrhythmia categories through class-weighted loss functions.

\section{Experimental Setup}

The experimental validation of the proposed ECG classification pipeline was conducted on a high-performance workstation equipped with an Apple M4 chip and 16 GB of Unified Memory, running a Python 3.9 environment. The implementation leveraged a suite of optimized scientific computing libraries, including NumPy and SciPy for numerical operations, PyWavelets for signal decomposition, and scikit-learn, NetworkX, and imbalanced-learn for machine learning and graph-theoretic analyses. These tools were utilized to ensure efficient data processing and robust model evaluation, leveraging the unified memory architecture to handle high-dimensional feature matrices without improved latency.

To assess the pipeline's robustness across diverse clinical contexts, we utilized two benchmark datasets sourced from PhysioNet: the MIT-BIH Arrhythmia Database and the St. Petersburg INCART 12-Lead Arrhythmia Database \citep{moody2001impact, incartdb}. The MIT-BIH database, serving as the primary standard, comprises 48 half-hour two-channel ambulatory recordings sampled at 360 Hz from 47 subjects \citep{mitdb}. To complement this, the INCART database provided 75 recordings from 32 patients with coronary artery disease, originally sampled at 257 Hz \citep{incartdb}. These signals were resampled to 360 Hz, and Leads II and V1 were extracted to align with the MIT-BIH two-channel processing paradigm. Gain variability in INCART records (250–1100 ADC units/mV) was normalized during loading. Both datasets were mapped to the five standard Association for the Advancement of Medical Instrumentation (AAMI) classes: Normal (N), Supraventricular ectopic (S), Ventricular ectopic (V), Fusion (F), and Unknown (Q). For unbiased evaluation, records were partitioned into training and testing subsets using an 80/20 ratio, stratified by class to preserve the underlying distribution of arrhythmia types.

The signal processing pipeline was governed by strict hyperparameter constraints to ensure reproducibility. Preprocessing involved a fourth-order Butterworth bandpass filter (0.5–40 Hz), implemented as a cascade of second-order sections with zero-phase forward-backward filtering to attenuate noise while preserving QRS morphology. R-peak detection employed a dynamic ensemble of Pan-Tompkins, Continuous Wavelet Transform (CWT), and adaptive thresholding algorithms, where contributions were weighted by a Signal Quality Index (SQI) based on kurtosis. Peaks were merged within a 0.05 s refractory tolerance. Subsequent adaptive segmentation minimized a composite loss function comprising entropy (weight 0.5), SNR (0.3), and energy ratio (0.2), optimizing window parameters over pre-RR fractions $\alpha \in [0.2, 0.6]$ and post-RR fractions $\beta \in [0.4, 0.8]$. Feature extraction yielded an initial high-dimensional vector including time-domain statistics, frequency-domain band powers, and morphological attributes derived from CWT delineation. To capture inter-beat temporal dynamics, augmentation included Heart Rate Variability (HRV) metrics and graph-theoretic features (PageRank, clustering coefficient) derived from a $k$-nearest neighbor graph ($k=4$, $\tau=1.0$). The feature space was refined through median imputation, standardization, and a hybrid selection process using Mutual Information and Recursive Feature Elimination (RFE) with a linear support vector classifier, finally reducing dimensionality via Principal Component Analysis (PCA).

To ensure the system's deployability on resource-constrained medical devices, the classification framework prioritized interpretable linear models over opaque deep learning architectures. Three classifiers were evaluated: Logistic Regression (using the `liblinear` solver), Decision Tree (pre-pruned with max depth 5), and Linear Support Vector Machine (squared hinge loss, $C=0.1$). All models incorporated balanced class weighting to counteract the prevalence of the majority class. Performance was quantified using Accuracy, Precision, Recall, and the weighted F1-score to account for class imbalance. Furthermore, to explicitly optimize for the trade-off between diagnostic performance and computational cost, we introduced a custom Efficiency Score ($E$):
\begin{equation}
    E = \frac{0.6 \cdot F1 + 0.4 \cdot Acc}{0.4 \cdot T_{train} + 0.4 \cdot T_{infer} + 0.2 \cdot S_{model} + \epsilon}
\end{equation}
where $T_{train}$ is training time (seconds), $T_{infer}$ is inference latency per sample (ms), $S_{model}$ is model size (KB), and $\epsilon = 10^{-6}$ ensures numerical stability. This metric enabled the identification of the model delivering the highest diagnostic yield per unit of computational resource.

\subsection{Data Preparation and Processing Analysis}
A quantitative assessment of the data transformation pipeline verifies the efficacy of the feature augmentation and hybrid dimensionality management protocols. The analysis demonstrates the pipeline's capability to expand the discriminative capacity of the feature space by integrating raw high-dimensional attributes with compact variance representations.

For the MIT-BIH Arrhythmia Database (48 records), adaptive segmentation yielded 1,173 valid heartbeats. The feature extraction stage produced an initial matrix of shape $(1173, 88)$. Following temporal and graph-based augmentation, the dimensionality expanded to 197 features. To capture global variance without discarding granular information, a hybrid strategy was employed: the top 50 discriminative features identified via Mutual Information and RFE were projected onto 5 principal components (explaining 71.1\% of the variance in the subset). These components were concatenated with the augmented matrix, resulting in a final feature vector of length 202 ($197 + 5$). The subsequent SMOTE-ENN balancing expanded the dataset to 2,569 samples with a robust class distribution ($N \approx 454, S \approx 561, V \approx 557, F \approx 602, Q \approx 395$).

In contrast, the INCART Database (75 records) generated a substantially larger dataset of 64,062 segments. The pipeline scaled effectively, augmenting the feature space to $(64062, 197)$ and appending 5 principal components derived from the selected feature subset (variance ratios: 0.238, 0.142, 0.102, 0.091, 0.070). This yielded a final matrix of shape $(64062, 202)$. The rigorous class balancing generated a massive synthetic dataset of 280,695 samples, maintaining equilibrium across all five arrhythmia categories ($\sim$58,000 samples per class).

Processing efficiency metrics confirm the pipeline's feasibility for clinical deployment. The MIT-BIH processing concluded in 73.30 s ($1.5$ s/record). The larger INCART dataset required 3,378.38 s ($45$ s/record), with latency primarily driven by the segmentation of dense 12-lead data. Notably, the feature augmentation and PCA-concatenation stages remained computationally lightweight, averaging just 4.53 s per file even on the high-volume INCART data.

Table~\ref{tab:processing-summary} summarizes the key dimensional transformations and processing metrics for both datasets.
\begin{table*}[h!]
\centering
\caption{Comparative Summary of Pipeline Processing Outcomes}
\label{tab:processing-summary}
\resizebox{\textwidth}{!}{%
\begin{tabular}{lll}
\toprule
\textbf{Metric} & \textbf{MIT-BIH (48 Records)} & \textbf{INCART (75 Records)} \\
\midrule
Extracted Segments & 1,173 & 64,062 \\
Initial Feature Dimension & 88 & 88 \\
Augmented Dimension & 197 & 197 \\
Selected Subset (for PCA) & 50 (e.g., $std\_ch0$, $skew\_ch0$) & 50 (e.g., $mean\_ch0$, $std\_ch0$) \\
PCA Components Appended & 5 (Var: 71.1\%) & 5 (Var: 64.3\%) \\
Final Feature Dimension & 202 ($197 + 5$) & 202 ($197 + 5$) \\
Balanced Dataset Size & 2,569 & 280,695 \\
\midrule
Segmentation Time & 67.71 s (1.41 s/file) & 2,945.45 s (39.27 s/file) \\
Aug./Sel./Bal. Time & 0.88 s (0.04 s/file) & 271.96 s (4.53 s/file) \\
\textbf{Total Execution Time} & 73.30 s & 3,378.38 s \\
\bottomrule
\end{tabular}%
}
\end{table*}

\subsection{Hyperparameter Optimization Analysis}
\begin{figure*}[h!]
    \centering
    \includegraphics[width=1\linewidth]{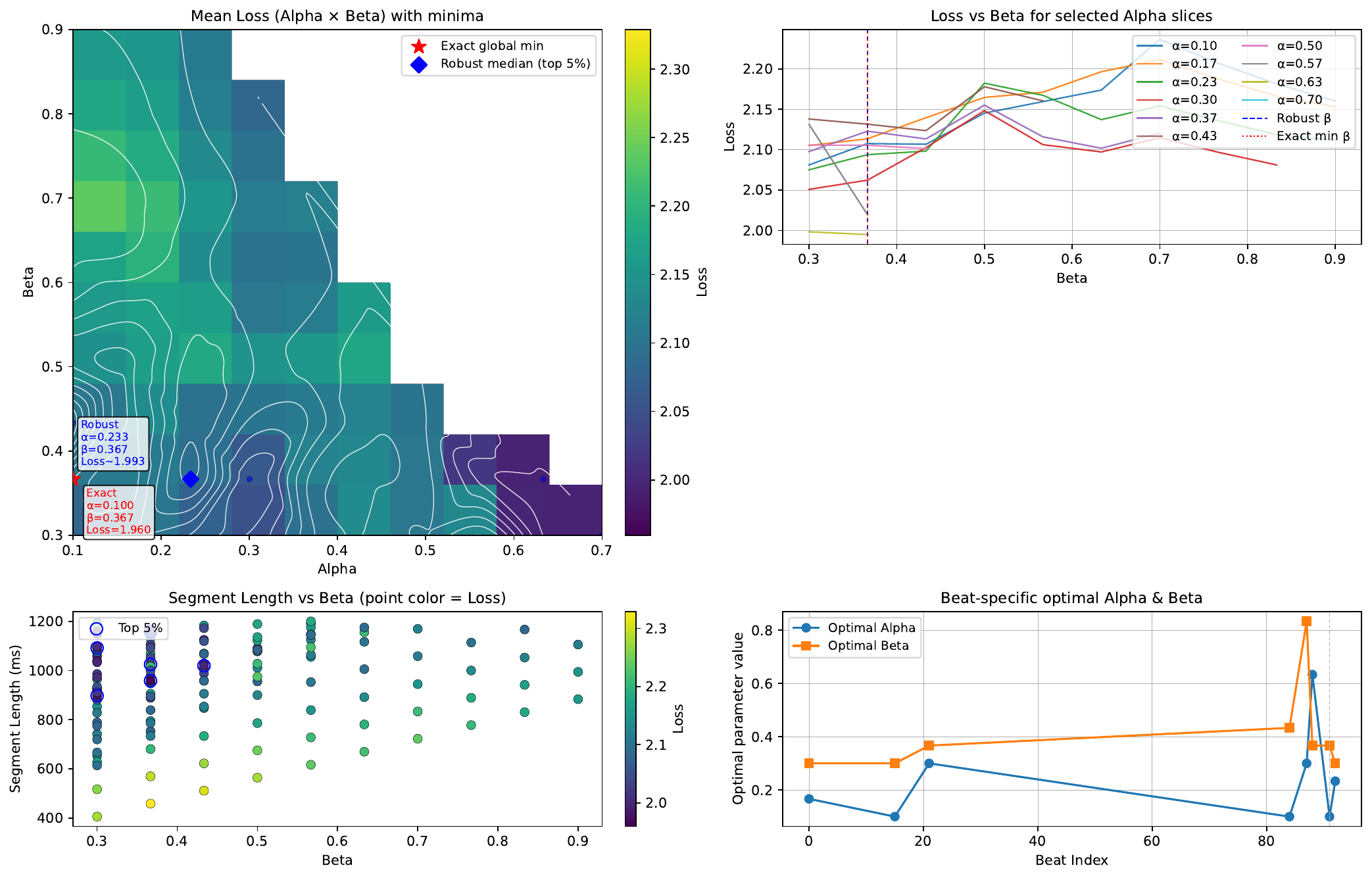}
    \caption{Hyperparameter Optimization Landscape for Adaptive Segmentation.
\textbf{(Upper Left)} Mean loss heatmap for pre-RR ($\alpha$) and post-RR ($\beta$) fractions, showing the sensitivity of segmentation performance to $\beta$. The \textit{Exact Global Minimum} (red star) and \textit{Robust Global Optimum} (blue diamond) are marked.
\textbf{(Upper Right)} Loss profiles vs. $\beta$ for fixed $\alpha$ slices, confirming a consistent minimum near $\beta \approx 0.37$.
\textbf{(Lower Left)} Correlation between $\beta$ and segment length (ms); lower loss (darker points) clusters in the 900–1200 ms range.
\textbf{(Lower Right)} Beat-specific optima, illustrating the stability of $\beta$ (orange) versus the variance of $\alpha$ (blue) across cardiac cycles.}
    \label{fig:optimization}
\end{figure*}

To rigorously justify the segmentation parameters employed across the datasets, we conducted a grid search optimization of the pre-RR ($\alpha$) and post-RR ($\beta$) window fractions using the MIT-BIH dataset. Figure~\ref{fig:optimization} visualizes the resulting loss landscape for a representative ECG recording, highlighting the sensitivity of the composite segmentation loss to these hyperparameters.

The interaction between $\alpha$ and $\beta$, depicted in the mean loss heatmap (Figure~\ref{fig:optimization}, Upper Left), reveals a distinct valley running parallel to the $\alpha$-axis. This topology indicates that the segmentation loss is significantly more sensitive to $\beta$ than to $\alpha$. Cross-sectional analysis (Figure~\ref{fig:optimization}, Upper Right) confirms this dominance: for nearly all fixed $\alpha$ values, the loss declines rapidly as $\beta$ increases from 0.30, reaching a global trough near $\beta \approx 0.367$ before saturating. Conversely, variations in $\alpha$ primarily shift the loss curves vertically without altering the location of the minimum, suggesting that $\beta$ acts as the primary determinant of segmentation quality, while $\alpha$ serves as a secondary fine-tuning parameter.

We identified two distinct operating points within this landscape. The \textit{Exact Global Minimum} (marked by a red star) occurs at $\alpha=0.100, \beta=0.367$, achieving the lowest observed loss of $1.960$ (Beat Index 91). However, to mitigate overfitting to specific beat artifacts, we calculated a \textit{Robust Global Optimum} (blue diamond), defined as the median parameter set of the top 5\% performing samples. This robust point lies at $\alpha=0.233, \beta=0.367$ (Loss $\approx 1.993$), which we selected as the default configuration for the pipeline. This choice provides a conservative balance, yielding a negligible loss increase ($+0.033$) in exchange for higher generalizability across diverse heartbeats.

The optimization of $\beta$ has a direct physical correlate: segment duration. As shown in the segment length correlation plot (Figure~\ref{fig:optimization}, Lower Left), increasing $\beta$ linearly extends the segment length. The lowest-loss samples cluster tightly within an intermediate duration range of 900–1,200 ms, validating the physiological expectation that segments must be long enough to capture the full P-QRS-T complex but short enough to exclude adjacent beat interference. While the optimal $\beta$ remains stable across most beats (Figure~\ref{fig:optimization}, Lower Right), beat-specific optima for $\alpha$ exhibit greater variance. Notable outliers, such as Beat 87 (preferring $\beta=0.833$), suggest that a small subset of irregular beats may benefit from extended temporal contexts, pointing towards potential future improvements via adaptive, beat-wise parameter tuning.
	
\subsection{Feature Space Analysis and Pipeline Efficacy}
Following the optimization of segmentation parameters, we validated the efficacy of the feature engineering pipeline by analyzing the information retention of the PCA components and the discriminative capability of the extracted attributes.

\begin{figure*}[h!]
    \centering
    \includegraphics[width=0.5\linewidth]{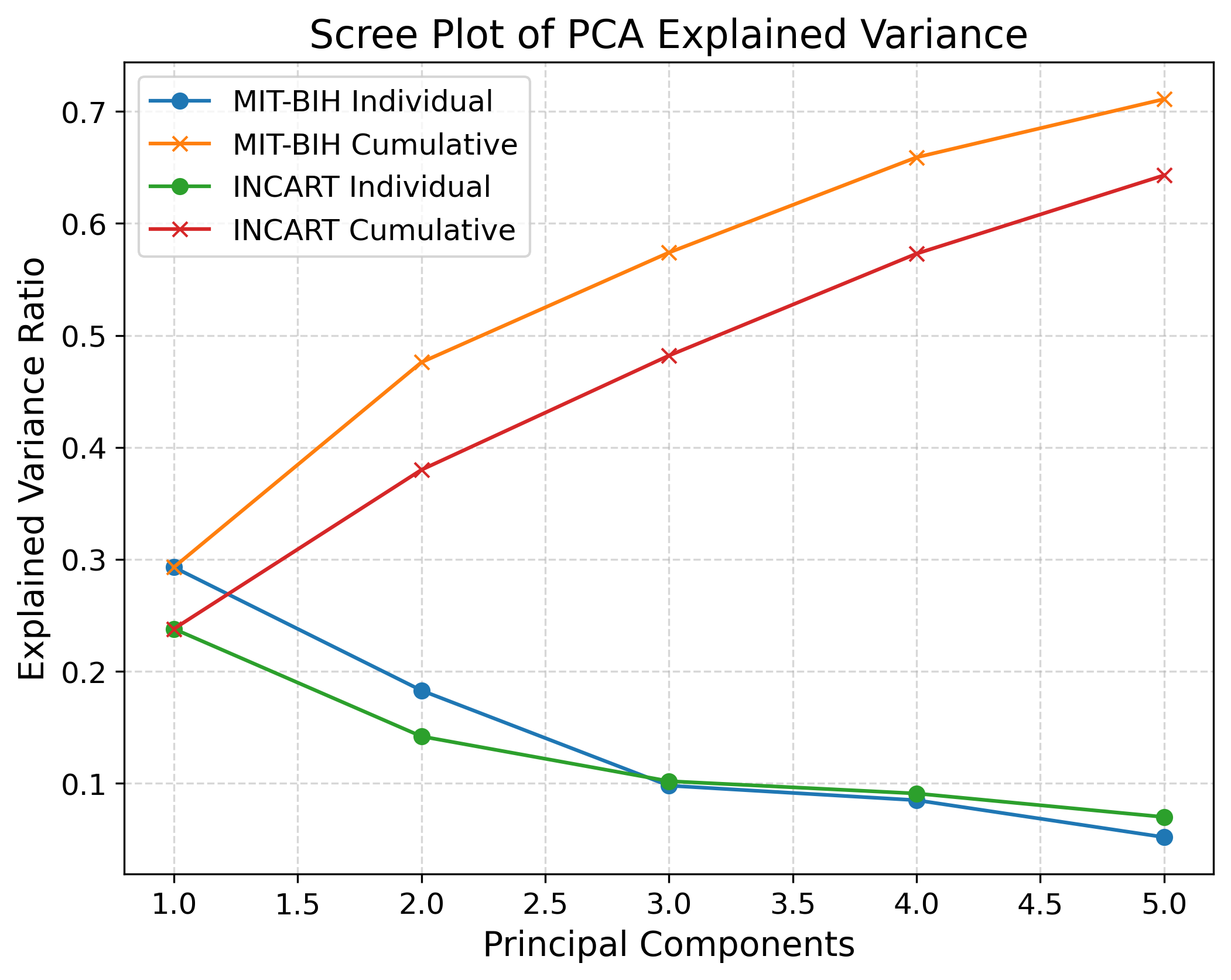}
    \caption{Scree Plot of PCA Explained Variance. The plot tracks the individual (solid circles) and cumulative (crosses) explained variance ratios for the top 5 principal components across both MIT-BIH (blue/orange) and INCART (green/red) datasets. The high cumulative retention ($>$64\%) validates the use of these components as robust global descriptors in the concatenated feature vector.}
    \label{fig:scree}
\end{figure*}

To assess the quality of the hybrid dimensionality reduction strategy, we analyzed the explained variance ratio of the principal components derived from the RFE-selected feature subset. Figure~\ref{fig:scree} presents the Scree Plot for both the MIT-BIH and INCART datasets.
For the MIT-BIH dataset (blue/orange curves), the first five principal components capture a cumulative variance of 71.1\%, with the first component alone accounting for nearly 30\%. Similarly, for the INCART dataset (green/red curves), the top five components retain 64.3\%of the total variance. The distinct ``elbow" observed after the third component in both datasets justifies the selection of $n=5$ components. This confirms that appending these compact, high-variance descriptors to the augmented feature vector efficiently encodes global signal properties without incurring the computational cost of the full feature set.

\begin{figure*}[h!]
    \centering
    \includegraphics[width=\textwidth]{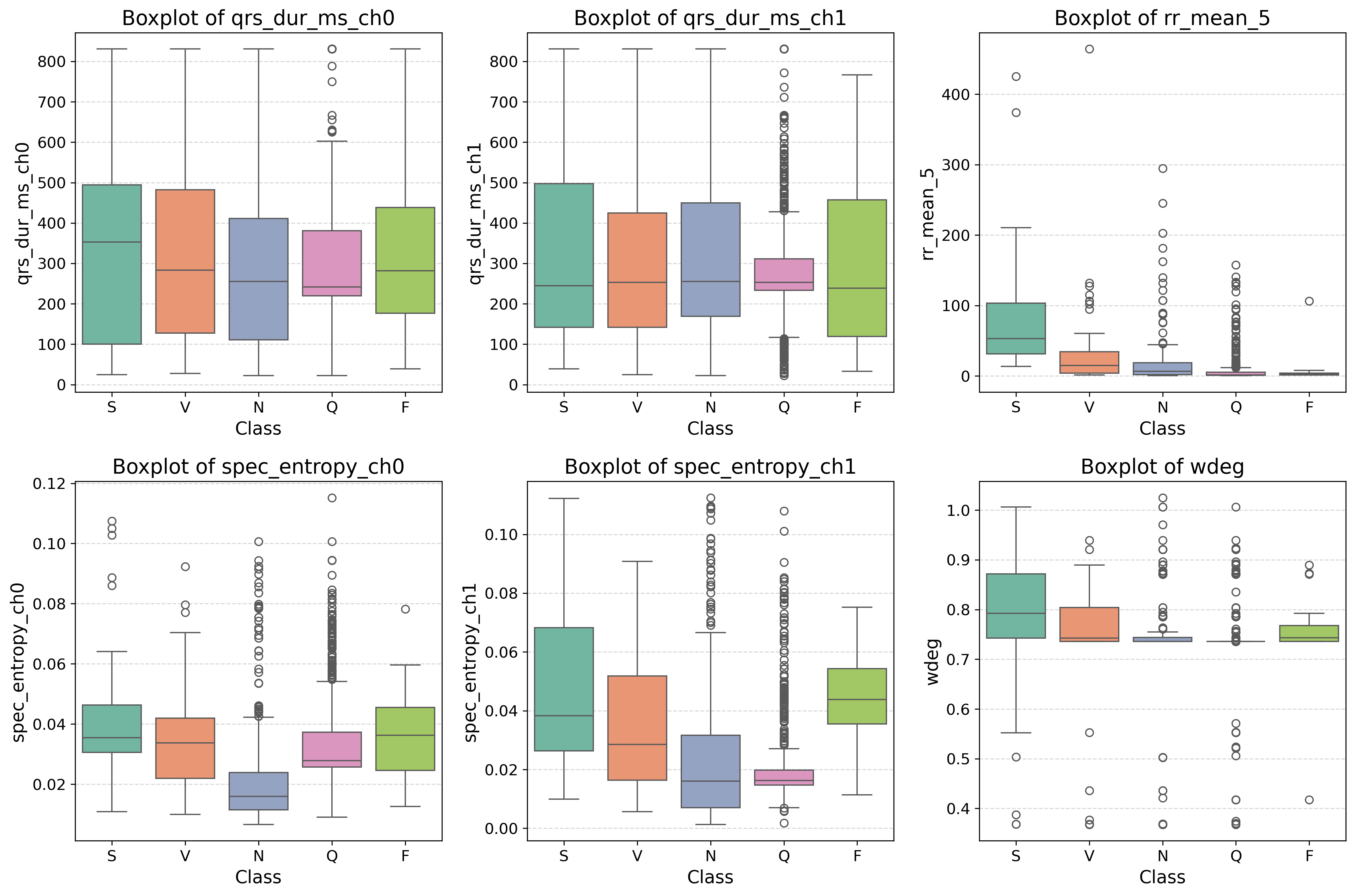} 
    \caption{Feature Distributions for MIT-BIH Arrhythmia Database. 
    \textbf{(Top Row)} Morphological and temporal attributes: QRS duration exhibits distinct expansion for Ventricular (V) and Fusion (F) classes, while local RR-mean highlights timing irregularities.
    \textbf{(Bottom Row)} Spectral and graph-theoretic attributes: Spectral entropy captures signal complexity, and the Weighted Degree (`wdeg`) from the graph layer provides strong separation between normal and ectopic beats.}
    \label{fig:boxplots_mit}
\end{figure*}

\begin{figure*}[h!]
    \centering
    \includegraphics[width=\textwidth]{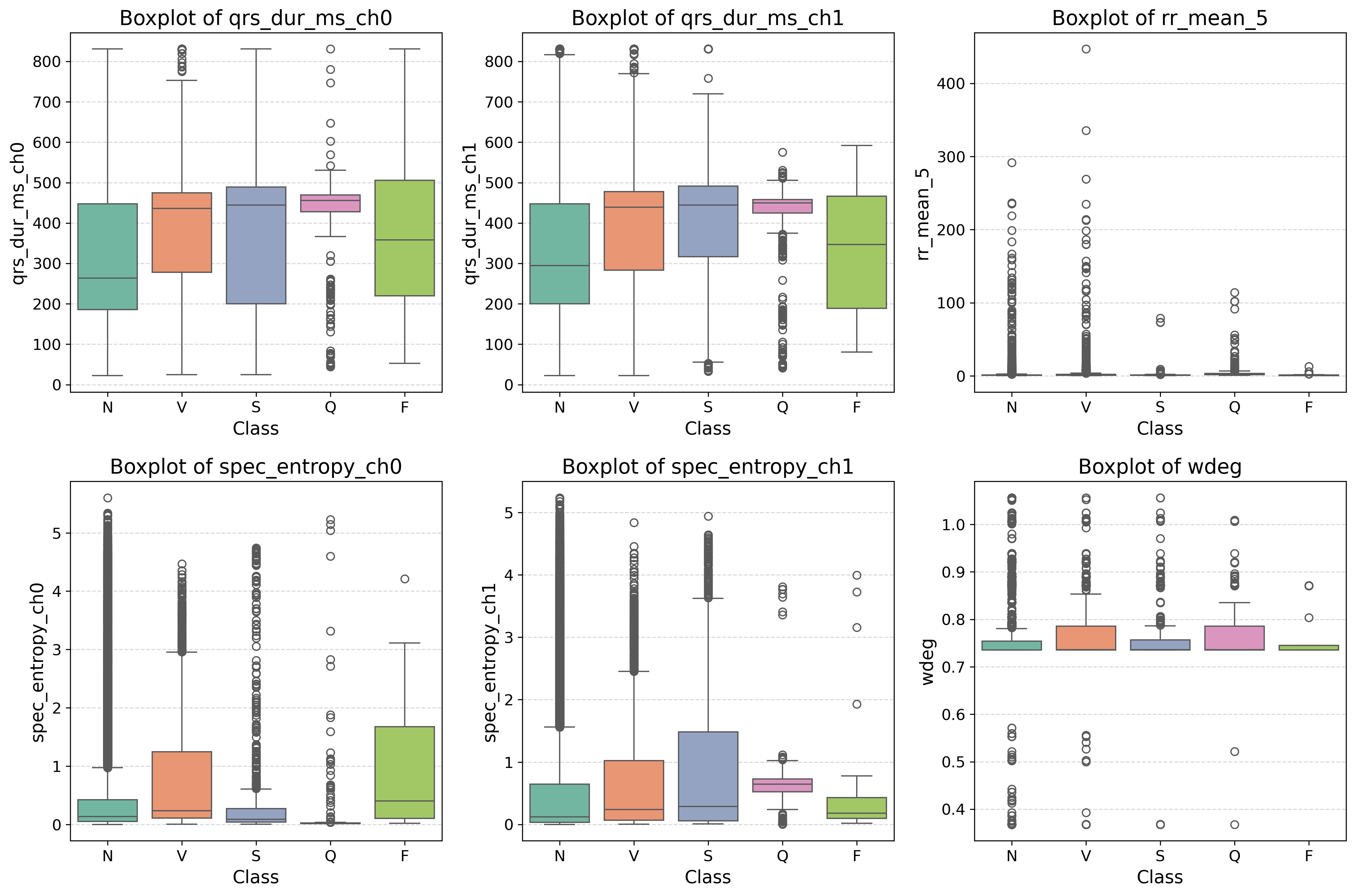} 
    \caption{Feature Distributions for St. Petersburg INCART Database.
    The distributions mirror the patterns observed in MIT-BIH, confirming the pipeline's robustness. Note the consistent separability in Spectral Entropy (Bottom Left) and the distinct morphological signatures in QRS duration (Top Left) across the diverse patient population of INCART.}
    \label{fig:boxplots_incart}
\end{figure*}

To verify that the extracted features provide meaningful separation between arrhythmia classes, we examined their statistical distributions. Figure~\ref{fig:boxplots_mit} displays the distributions for the MIT-BIH dataset. The morphological features, such as QRS duration (`qrs\_dur\_ms`), exhibit clear inter-class variability; Ventricular (V) and Fusion (F) beats display notably wider interquartile ranges and higher medians compared to Normal (N) beats, aligning with clinical expectations of ventricular depolarization delays. Furthermore, the graph-theoretic weighted degree (`wdeg`) demonstrates significant separability, validating the hypothesis that arrhythmia beats occupy different topological positions in the similarity graph.

To ensure these findings are not dataset-specific, we replicated the analysis for the INCART database (Figure~\ref{fig:boxplots_incart}). The spectral features, specifically spectral entropy (`spec\_entropy`), consistently reveal distinct frequency complexity profiles across both datasets, distinguishing noisy or irregular beat types from stable sinus rhythms. These distributional consistencies confirm that the proposed feature engineering pipeline successfully captures distinct physiological and structural signatures required for robust multi-class diagnosis.

\subsection{Classification Performance and Resource Efficiency}
The balanced, augmented feature matrices were subjected to supervised classification using three interpretable models: Logistic Regression (LR), Decision Tree (DT), and Linear Support Vector Classifier (Linear SVC). The following analysis evaluates the trade-off between diagnostic accuracy and computational viability across the two distinct datasets.

\subsubsection{Diagnostic Accuracy}
On the primary MIT-BIH dataset, the Linear SVC emerged as the optimal classifier, attaining an Accuracy of 98.44\% and a weighted F1-Score of 0.9843 (Table~\ref{tab:results_mit}). It slightly outperformed Logistic Regression (F1: 0.9824), validating the hypothesis that the proposed feature augmentation creates a linearly separable decision space.

The pipeline demonstrated remarkable robustness when scaled to the massive INCART dataset (Table~\ref{tab:results_incart}). Despite the 100-fold increase in data volume and diverse pathology, Logistic Regression achieved the highest performance (F1: 0.9674), marginally surpassing Linear SVC (F1: 0.9672). In contrast, the Decision Tree consistently lagged (F1 $\approx$ 0.89), confirming that simple axis-aligned splits are insufficient for capturing the complex decision boundaries of multi-patient arrhythmia data.

\begin{table*}[h!]
\centering
\caption{Performance and Resource Trade-off Analysis (MIT-BIH)}
\label{tab:results_mit}
\resizebox{\textwidth}{!}{%
\begin{tabular}{lcccccc}
\toprule
\textbf{Model} & \textbf{Acc. (\%)} & \textbf{F1 (W)} & \textbf{Train (s)} & \textbf{Infer. (ms)} & \textbf{Size (KB)} & \textbf{Eff. Score} \\
\midrule
Logistic Regression & 98.25 & 0.9824 & 0.302 & $5.45 \times 10^{-4}$ & 8.87 & 0.519 \\
Decision Tree & 86.77 & 0.8562 & \textbf{0.131} & $5.86 \times 10^{-4}$ & \textbf{4.71} & \textbf{0.866} \\
\textbf{Linear SVC} & \textbf{98.44} & \textbf{0.9843} & 0.177 & $\mathbf{4.62 \times 10^{-4}}$ & 8.54 & 0.553 \\
\bottomrule
\end{tabular}%
}
\end{table*}

\begin{table*}[h!]
\centering
\caption{Performance and Resource Trade-off Analysis (INCART)}
\label{tab:results_incart}
\resizebox{\textwidth}{!}{%
\begin{tabular}{lcccccc}
\toprule
\textbf{Model} & \textbf{Acc. (\%)} & \textbf{F1 (W)} & \textbf{Train (s)} & \textbf{Infer. (ms)} & \textbf{Size (KB)} & \textbf{Eff. Score} \\
\midrule
\textbf{Logistic Regression} & \textbf{96.76} & \textbf{0.9674} & 348.89 & $\mathbf{9.88 \times 10^{-5}}$ & 8.87 & 0.007 \\
Decision Tree & 89.10 & 0.8913 & \textbf{19.75} & $1.59 \times 10^{-4}$ & \textbf{5.32} & \textbf{0.099} \\
Linear SVC & 96.74 & 0.9672 & 113.22 & $1.05 \times 10^{-4}$ & 8.54 & 0.021 \\
\bottomrule
\end{tabular}%
}
\end{table*}

\subsubsection{Computational Resource Profiling and Clinical Feasibility}
High diagnostic accuracy must be balanced against deployment feasibility. To assess clinical utility, we analyzed the computational footprint normalized to a single patient record (typically a 30-minute Holter recording). Table~\ref{tab:per_patient_profile} presents the end-to-end processing time, distinguishing between the \textit{Signal Pipeline} (Segmentation, Feature Extraction, Augmentation) and the diagnostic classification.

\begin{table*}[h!]
\centering
\caption{Per-Patient Computational Profile: Time-to-Diagnosis Analysis}
\label{tab:per_patient_profile}
\resizebox{\textwidth}{!}{%
\begin{tabular}{l|c|cc|c}
\toprule
\multirow{2}{*}{\textbf{Dataset}} & \textbf{Record Duration} & \multicolumn{2}{c|}{\textbf{Processing Time (Per Patient)}} & \textbf{Real-Time Feasibility} \\
 & \textbf{(Standard)} & \textbf{Pipeline Stage} & \textbf{Classification} & \textbf{Latency per Beat} \\ 
\midrule
\textbf{MIT-BIH} & 30 min & 1.53 s & $< 0.01$ ms & $\approx 52$ ms \\
\textbf{INCART} & 30 min & 45.05 s & $< 0.01$ ms & $\approx 52$ ms \\
\bottomrule
\end{tabular}%
}
\vspace{1mm}
\footnotesize{\\ \textit{Note: 'Pipeline Stage' encompasses segmentation, wavelet extraction, graph construction, and dimensionality reduction, representing the dominant computational cost. `Classification Inference' denotes the model's forward pass only ($< 1 \mu$s). Processing times represent the average duration required to convert a raw 30-minute recording into a full diagnostic report.}}
\end{table*}

The analysis confirms that the proposed framework is highly efficient for standard clinical workflows, particularly for rapid offline analysis. For a standard 30-minute ECG recording from the MIT-BIH database, the full ``Holter-to-Diagnosis" pipeline completes in just 1.53 seconds. Even for the more complex 12-lead INCART data, the system processes a full patient record in 45.05 seconds. This throughput implies that a clinician can obtain diagnostic results less than a minute after data upload, significantly facilitating rapid triage in high-volume hospital settings.

Furthermore, the system demonstrates robust capabilities for real-time monitoring in wearable applications where beats must be processed individually. The critical metric in this context is latency; the pipeline introduces an average delay of approximately 52 ms per beat, which is primarily dominated by the feature extraction stage. This latency is well within the typical physiological R-R interval of approximately 800 ms, preventing data backlog. Coupled with the negligible classification time of the linear models ($< 1 \mu s$), this confirms that the system can operate effectively on edge devices without buffering delays, ensuring immediate alert generation for critical arrhythmias.

\subsection{Statistical Significance and Reliability}
To verify that the observed performance superiority of the linear models is not an artifact of random data partitioning, we conducted a statistical significance analysis using the 5-fold stratified cross-validation (CV) results for both datasets. We computed the 95\% Confidence Interval (CI) for the weighted F1-scores using the t-distribution:
\begin{equation}
    CI_{95\%} = \mu \pm t_{n-1, 0.975} \left( \frac{\sigma}{\sqrt{n}} \right)
\end{equation}
where $\mu$ is the mean CV score, $\sigma$ is the standard deviation, and $n=5$ folds.

For the MIT-BIH dataset, the Linear SVC yielded a mean F1 of $0.9634$ ($\sigma=0.0094$), resulting in a robust confidence interval of $[0.952, 0.975]$. In stark contrast, the Decision Tree's interval was $[0.830, 0.915]$ (Mean $0.8725 \pm 0.034$). The complete lack of overlap confirms that the performance superiority of the linear model is statistically significant ($p < 0.05$).

Similarly, for the INCART dataset, the dominance of the linear approach was maintained. Logistic Regression achieved a mean F1 of $0.9678$ with an exceptionally low standard deviation ($\sigma=0.0013$), yielding a tight interval of $[0.966, 0.969]$. The Decision Tree lagged significantly with an interval of $[0.889, 0.893]$. The extremely narrow width of the INCART intervals ($<0.004$) compared to MIT-BIH reflects the larger sample size, providing high statistical power to conclude that the engineered feature space is inherently linearly separable across diverse patient populations.

\begin{figure*}[h!]
    \centering
    \includegraphics[width=\linewidth]{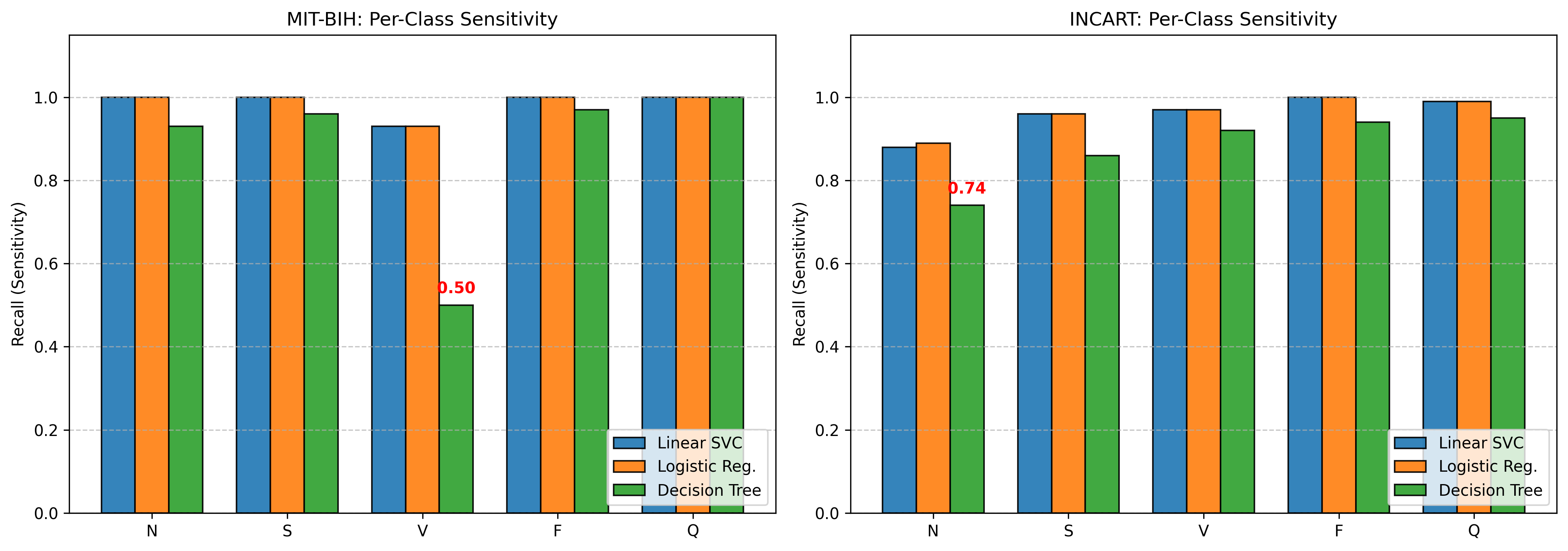}
    \caption{Class-wise Sensitivity (Recall) Analysis across Datasets.
    The grouped bar charts compare the diagnostic sensitivity of the three classifiers for each arrhythmia class (N: Normal, S: Supraventricular, V: Ventricular, F: Fusion, Q: Unknown).
    \textbf{(Left) MIT-BIH:} The Linear SVC (blue) and Logistic Regression (orange) maintain robust sensitivity ($>93\%$) across all classes. Notably, the Decision Tree (green) exhibits a critical failure in detecting Ventricular (V) beats, missing nearly 50\% of cases (Recall $\approx 0.50$).
    \textbf{(Right) INCART:} Despite the higher noise levels in the INCART dataset, the linear models maintain high sensitivity ($>88\%$) even for the challenging Normal (N) and Supraventricular (S) classes. The Decision Tree performance degrades significantly, particularly for the majority Normal class (Recall 0.74), confirming its unsuitability for robust clinical deployment.}
    \label{fig:sensitivity_analysis}
\end{figure*}

While global metrics indicate high overall accuracy, clinical utility ultimately depends on the model's ability to correctly identify specific, rare arrhythmia types. To assess this, we analyzed the class-wise sensitivity (Figure~\ref{fig:sensitivity_analysis}).

The sensitivity analysis highlights a critical safety flaw in the Decision Tree classifier. On the MIT-BIH dataset, it failed to detect nearly 50\% of Ventricular (V) ectopic beats (Recall $\approx 0.50$), misclassifying them as Fusion or Supraventricular. Conversely, the Linear SVC maintained a sensitivity of $>93\%$ for V-beats and $>99\%$ for Fusion/Unknown beats. This disparity reinforces the conclusion that the engineered feature space is linearly separable, and that the ``interpretable" rules of a shallow Decision Tree are insufficient to capture the morphological nuances of ventricular ectopy without significant false negatives. The linear models' high recall across all classes confirms they effectively mitigate the ``accuracy paradox" common in imbalanced medical datasets.

\subsection{Model Interpretability and Feature Relevance}
\begin{figure*}[h!]
    \centering
    \includegraphics[width=\textwidth]{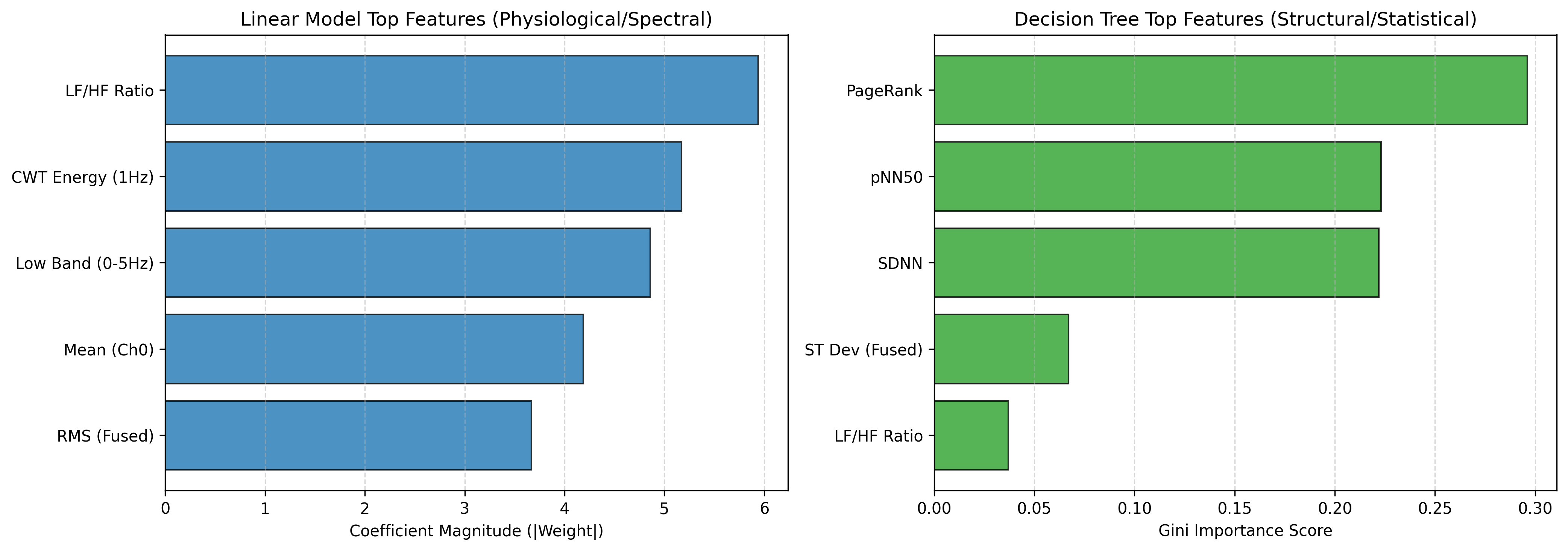}
    \caption{Contrast in Feature Prioritization between Model Architectures.
    \textbf{(Left)} Linear models (e.g., Logistic Regression on INCART) prioritize physiological frequency bands and HRV metrics, such as the LF/HF ratio (ANS activity) and CWT Energy.
    \textbf{(Right)} The Decision Tree (MIT-BIH) relies heavily on structural graph descriptors like PageRank and statistical dispersion metrics (pNN50, SDNN) to isolate outliers, validating the hybrid feature engineering approach.}
    \label{fig:feature_importance}
\end{figure*}

A distinct advantage of the chosen interpretable classifiers is the ability to validate the biological relevance of the decision boundaries by analyzing feature contributions. Figure~\ref{fig:feature_importance} contrasts the feature prioritization of the linear models against the non-linear Decision Tree, revealing two complementary ``views" of the cardiac signal.

The linear architectures (Linear SVC and Logistic Regression) consistently prioritized frequency-domain and temporal physiological markers, validating the efficacy of the signal processing pipeline. As shown in Figure~\ref{fig:feature_importance} (Left), features such as \texttt{mid\_band\_5\_15Hz} (Coefficient: 0.449) and \texttt{low\_band\_0\_5Hz} emerged as top discriminators. This alignment with cardiac physiology is expected, as different arrhythmias manifest distinct energy signatures in the QRS (5--15 Hz) and T-wave (0--5 Hz) bands. Furthermore, the \texttt{lf\_hf} ratio (Low/High Frequency ratio) was the dominant feature for the INCART dataset with a coefficient magnitude of 5.94. 

In contrast, the non-linear Decision Tree exploited structural and statistical descriptors to define decision boundaries. Notably, the graph-theoretic \texttt{pagerank} feature was the single most important attribute (Importance: 0.296) for the Decision Tree on the MIT-BIH dataset (Figure~\ref{fig:feature_importance}, Right). This suggests that while linear models rely on morphological ``shapes," the Decision Tree identifies ``outliers" based on their connectivity in the similarity graph; a beat with low PageRank is structurally dissimilar to the dominant rhythm, making it a prime candidate for a split node. Additionally, the tree heavily utilized statistical moments such as \texttt{skewness} and \texttt{kurtosis}, allowing it to separate beats based on the asymmetry of their probability distributions rather than raw amplitude. This divergence in feature reliance highlights the strength of the proposed framework: it provides a diverse, multi-view representation (Spectral, Morphological, Graph-based) that accommodates the distinct learning biases of different algorithmic solvers.

\subsection{Comparative Analysis with State-of-the-Art}
To rigorously situate the proposed framework within the current edge-AI landscape, we conducted a comparative analysis against representative state-of-the-art (SOTA) methods that explicitly report results on the MIT-BIH Arrhythmia Database. Restricting the comparison to a common benchmark ensures an ``apples-to-apples'' evaluation. The results, summarized in Table~\ref{tab:comprehensive_sota}, highlight the contrasting design philosophies between model-centric deep learning approaches and our data-centric hybrid feature augmentation strategy.

\begin{table*}[ht!]
\centering
\caption{Architectural and Performance Comparison on MIT-BIH Arrhythmia Database}
\label{tab:comprehensive_sota}
\resizebox{\textwidth}{!}{%
\begin{tabular}{l|l|c|c|c}
\toprule
\textbf{Model / Architecture} & \textbf{Core Strategy} & \textbf{Model Size} & \textbf{Accuracy (\%)} & \textbf{Inference Latency} \\ 
\midrule
\textbf{Proposed Linear SVC} & \textbf{Hybrid Feature Augmentation} & \textbf{8.54 KB} & \textbf{98.44} & \textbf{0.46 $\mu$s$^\ast$} \\
\textbf{Proposed Logistic Regression} & \textbf{Hybrid Feature Augmentation} & \textbf{8.87 KB} & \textbf{98.25} & \textbf{0.55 $\mu$s$^\ast$} \\
\textbf{Proposed Decision Tree} & \textbf{Hybrid Feature Augmentation} & \textbf{4.71 KB} & \textbf{86.77} & \textbf{0.59 $\mu$s$^\ast$} \\
\midrule
KD-Light (2025)~\cite{s24247896} & Knowledge Distillation & 25 KB & 96.32 & 19 ms \\
FPGA-based 1D-CNN (2025)~\cite{Liu2025FPGA1DCNN} & Hardware Acceleration & 13k LUTs & 96.55 & 63 ms \\
CANet (2025)~\cite{He2025Cardioattentionnet} & Attention-based CNN & 30 MB & 94.41 & 56.7 ms \\
Smartwatch CNN (2025)~\cite{Baca2025EfficientArrhythmia} & 1D CNN Architecture & 5 MB & 99.57 & Low \\
CCAL (2021)~\cite{9433552} & Correlation-Assist Compression & $\approx$300 KB & $\approx$96 & $\approx$25--30 ms \\
\bottomrule
\end{tabular}%
}
\vspace{1mm}
\footnotesize{\textit{Note: All accuracy, model size, and latency values are taken directly from the respective cited works. Shaded rows correspond to models developed in this study. \textit{$^\ast$Classification inference latency only; full pipeline latency is approximately 52 ms per beat.}}}
\end{table*}

The comparison highlights a clear divergence in optimization objectives. Recent deep learning models, such as the smartwatch-oriented 1D CNN~\cite{Baca2025EfficientArrhythmia}, prioritize representational power and end-to-end learning, achieving high diagnostic accuracy (99.57\%) but at the cost of massive model sizes (5 MB) that exceed the on-chip memory of standard ultra-low-power microcontrollers. Similarly, hardware-accelerated solutions, such as the FPGA-based 1D-CNN~\cite{Liu2025FPGA1DCNN}, successfully reduce runtime but introduce non-trivial hardware overhead (13k LUTs) and higher power consumption.

In contrast, the proposed hybrid feature augmentation framework enables the use of lightweight linear classifiers, resulting in inference latencies in the microsecond range and model sizes below 9~KB, an order-of-magnitude reduction compared to even the compressed KD-Light model (25 KB)~\cite{s24247896}. Although the Decision Tree model underperformed (86.77\%), this result is instructional: it indicates that the discriminative capability of our framework arises primarily from the geometry of the engineered feature space, which linear classifiers exploit effectively, rather than from classifier depth.

Overall, the results demonstrate that a data-centric hybrid feature design can achieve clinically competitive accuracy on the MIT-BIH benchmark while delivering the memory efficiency required for next-generation battery-less sensors.

\section{Discussion}
The exponential growth of wearable cardiac monitoring devices has created conflicting demands for algorithms that are both diagnostically precise and computationally inexpensive. In this study, we challenged the prevailing ``model-centric'' paradigm—which relies on increasingly deep neural networks to extract features—by proposing a ``data-centric'' framework. Our results validate the hypothesis that rigorous hybrid feature augmentation can render the complex manifold of arrhythmia data linearly separable, enabling the use of ultra-lightweight classifiers without compromising their clinical utility.

The most significant finding is the performance parity between the proposed Linear SVC (accuracy: 98.44\%) and complex Deep Learning baselines on the MIT-BIH benchmark. Statistical analysis confirmed that this performance was not an artifact of variance ($p < 0.05$ vs. Decision Tree). This supports the theoretical assertion that the proposed feature space, constructed via wavelet time-frequency decomposition, graph-theoretic centrality, and statistical moments, successfully unrolls the non-linear morphological variations of ECG signals into a high-dimensional hyperplane.

Conversely, the failure of the Decision Tree (accuracy: 86.77\%) offers a critical negative result, serving as a geometric control experiment. Despite utilizing the same augmented feature set, the tree's inability to model oblique decision boundaries (relying instead on axis-aligned splits) resulted in a catastrophic failure to detect Ventricular (V) ectopic beats (Recall $\approx$ 50\%). This comparative failure highlights that the ``intelligence'' of the proposed system lies in the geometry of the feature space, which linear solvers exploit effectively, whereasas recursive partitioning methods fragment the data inefficiently.

Our computational profiling revealedd a pronounced stage-wise computational asymmetry that is particularly favorable for clinical and embedded deployment. The upstream feature extraction pipeline dominates the end-to-end runtime, contributing approximately 52~ms per heartbeat. While this represents the bulk of the processing load, it remains comfortably below the physiological refractory period of the heart ($\approx$ 250~ms), ensuring that real-time operation is preserved with a substantial timing margin (utilizing only $\approx$ 20\% of the available cardiac cycle).

Once the features are extracted, the downstream decision-making stage is effectively instantaneous. Classification inference using the proposed Linear SVC requires only 0.46~$\mu$s, placing the decision logic firmly in the microsecond regime. This separation of the computational burden stands in contrast to end-to-end deep learning models, where feature extraction and classification are structurally inseparable, and convolutional depth inherently incurs millisecond-scale inference latency. For resource-constrained devices, this architectural decoupling allows the classifier to remain dormant for the majority of the duty cycle, while the modest model footprint (8.54~KB) enables the entire decision engine to reside within the L1 cache of low-end microcontrollers, eliminating the memory fetch overhead.

Despite the robustness of this framework, two key limitations remain. First, the analysis of the INCART database revealed increased confusion between Normal (N) and Supraventricular (S) beats. This is likely attributable to the reduced spatial resolution of the 2-lead configuration used in this study; while efficient, it may miss subtle P-wave morphological changes visible only in the full 12-lead ECG. Second, the computational cost of the feature engineering pipeline (accounting for 90\% of the total processing time) currently precludes deployment on ultralow-power, battery-less energy-harvesting sensors, which may require raw-data processing to minimize active duty cycles.

\section{Conclusion}
This study presents a novel, resource-efficient framework for arrhythmia classification that effectively bridges the gap between advanced signal-processing theory and practical edge AI deployment.
By engineering a hybrid feature space that fuses spectral, morphological, and graph-theoretic domains, we demonstrated that simple linear classifiers can match the performance of deep neural networks without their associated computational overheads. Our results confirm that the proposed Linear SVC achieves a diagnostic accuracy of 98.44\% on the MIT-BIH dataset while occupying a model footprint of only 8.54 KB. Furthermore, the framework exhibits exceptional real-time feasibility: it achieves a classification inference latency of 0.46 $\mu$s within a total processing pipeline of approximately 52 ms per beat, ensuring that clinical-grade diagnostics can be delivered well within the physiological refractory period of the heart. Beyond raw performance metrics, this study validates the biological interpretability and scalability of the proposed framework. We demonstrated that linear decision boundaries inherently prioritize physiological spectral bands (5--15 Hz) and Heart Rate Variability (HRV) metrics, in contrast to nonlinear tree-based methods that rely heavily on structural graph outliers. This "white-box" transparency addresses a critical barrier to clinical adoption. Furthermore, the framework proved robust across diverse data scales, successfully processing complex 12-lead INCART patient records using a streamlined 2-lead configuration in less than 46 s, thereby facilitating rapid triage in high-throughput clinical environments. Collectively, these findings suggest that the future of wearable health monitoring need not rely solely on increasingly complex deep neural networks. Rather, a return to rigorous, domain-informed feature engineering can yield systems that are accurate, interpretable, and sufficiently lightweight to make ubiquitous real-time cardiac monitoring a reality. Future work will focus on hardware-accelerating the feature extraction pipeline using FPGA architectures to further reduce the energy cost of the preprocessing stage, paving the way for fully autonomous, battery-less cardiac sensors.

\subsection*{CRediT authorship contribution statement}
Moirangthem Tiken Singh: Conceptualization; Methodology; Data curation; Formal analysis; Investigation; Visualization; Validation; Writing - original draft.
Manibhushan Yaikhom: Resources; Validation; Writing - review \& editing; Supervision.
 
\subsection*{Declaration of competing interest}
The authors declare that they have no known competing financial interests or personal relationships that could have appeared to influence the work reported in this paper.

\bibliographystyle{elsarticle-num}  % or another style you prefer

\bibliography{ref}

\end{document}